\documentclass[10pt,twocolumn,letterpaper]{article}

\usepackage[pagenumbers]{cvpr} % To force page numbers, e.g. for an arXiv version

\usepackage{svg}
\usepackage{algpseudocode} 
\usepackage{graphicx}
\usepackage{amsmath}
\usepackage{amssymb}
\usepackage{booktabs}

\usepackage{multirow}
\usepackage[linesnumbered,ruled,vlined]{algorithm2e}
\usepackage[pagebackref,breaklinks,colorlinks]{hyperref}

\usepackage[capitalize]{cleveref}
\crefname{section}{Sec.}{Secs.}
\Crefname{section}{Section}{Sections}
\Crefname{table}{Table}{Tables}
\crefname{table}{Tab.}{Tabs.}

\def\cvprPaperID{*****} % *** Enter the CVPR Paper ID here
\def\confName{CVPR}
\def\confYear{2023}

\begin{document}

\title{VirtualPainting: Addressing Sparsity with Virtual Points and Distance-Aware Data Augmentation for 3D Object Detection}

%%%%%%%%% AUTHORS - PLEASE UPDATE
\author{Sudip Dhakal\\
University of North Texas\\
Denton, Texas, USA\\
{\tt\small}
\and
Dominic Carrillo, Deyuan Qu, Michael Nutt, Qing Yang, Song Fu\\
University of North Texas\\
Denton, Texas, USA\\
{\tt\small}
}

\maketitle

%%%%%%%%% ABSTRACT
\begin{abstract}
In recent times, there has been a notable surge in multimodal approaches that decorates raw LiDAR point clouds with camera-derived features to improve object detection performance. However, we found that these methods still grapple with the inherent sparsity of LiDAR point cloud data, primarily because fewer points are enriched with camera-derived features for sparsely distributed objects. We present an innovative approach that involves the generation of virtual LiDAR points using camera images and enhancing these virtual points with semantic labels obtained from image-based segmentation networks to tackle this issue and facilitate the detection of sparsely distributed objects, particularly those that are occluded or distant. Furthermore, we integrate a distance aware data augmentation (DADA) technique to enhance the model’s capability to recognize these sparsely distributed objects by generating specialized training samples. Our approach offers a versatile solution that can be seamlessly integrated into various 3D frameworks and 2D semantic segmentation methods, resulting in significantly improved overall detection accuracy. Evaluation on the KITTI and nuScenes datasets demonstrates substantial enhancements in both 3D and bird’s eye view (BEV) detection benchmarks.
\end{abstract}

%%%%%%%%% BODY TEXT
\section{Introduction}
\label{sec:intro}
\begin{figure}[t]
  \centering
  \includegraphics[width=1.0\linewidth]{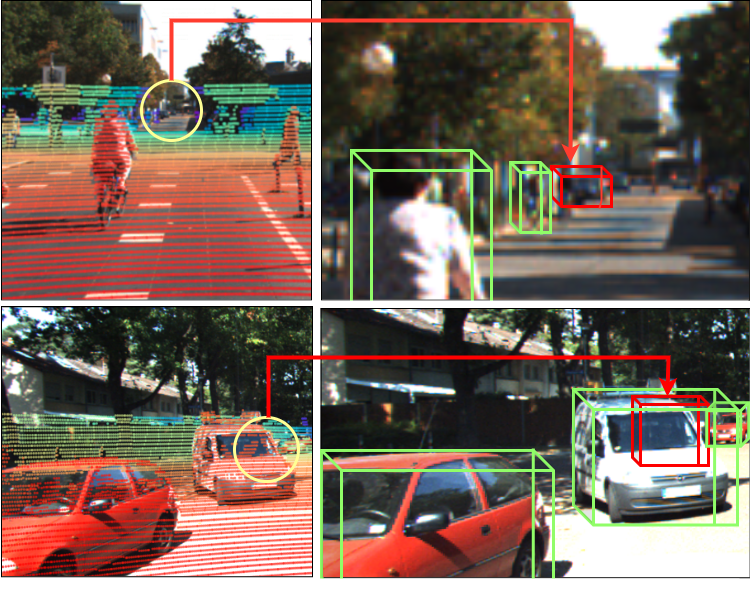}
  \caption{Drawback of painting based methods\cite{pointpainting}. The yellow circle serves as an indicator of the absence of point clouds in image projection, while the red bounding box signifies the consequent detection failure resulting from the sparse point cloud. Despite the presence of certain semantic cues for the object within the yellow circle, the absence of LiDAR points prevents the integration of these semantic cues with LiDAR data.}
  
\end{figure}
3D object detection plays a pivotal role in enhancing scene understanding for safe autonomous driving. In recent years, a large number of 3D object detection techniques have been implemented \cite{voxelnet, second, mv3d, 3dcvf, voxelrcnn, pointaugmenting, pointpainting}. These algorithms primarily leverage information from  LiDAR and camera sensors to perceive their surroundings. LiDAR provides low-resolution shape and depth information \cite{oasd}, while cameras capture dense images rich in color and textures. While multimodal based 3D object detection has made significant advancements recently, their performance still notably deteriorates when dealing with the sparse point cloud data. Recently, painting-based methods like PointPainting \cite{pointpainting}  have gained popularity in an attempt to address this issue. These methods decorate the LiDAR points with camera features. However, a fundamental challenge still persists in the case of objects lacking corresponding point clouds, such as distant and occluded objects. Despite the presence of camera features for such objects, there is no associated point cloud to complement these features. As a result, although they are beneficial for improving overall 3D detection performance, they still contend with sparse point clouds, as illustrated in Figure 1. 

\subsection{Limitations of Prior Art}
In response to the inherent sparsity of LiDAR data, several methods have been introduced to generate pseudo or virtual points. These methods bolster the sparse point clouds by introducing supplementary points around the existing LiDAR points. For example, MVP \cite{mvp_multimodal} generates virtual points by completing the depth information of 2D instance points using data from the nearest 3D points. Similarly, SFD \cite{9880017} creates virtual points based on depth completion networks \cite{10105781}. These virtual points play a crucial role in enhancing the geometric representation of distant objects, showcasing their potential to significantly enhance the performance of 3D detection methods. However, current implementations have yet to fully harness the advantages of integrating semantic results from semantic networks with virtual points. The incorporation of semantic information into the augmented point cloud, which includes both the original and virtual points, not only enriches the dataset but also increases the model's overall robustness.

Most recent fusion-based methods primarily concentrate on different fusion stages: early fusion, which involves combining LiDAR and camera data at an early stage; deep fusion, where features from both camera and LiDAR are combined by feature fusion; and late fusion, which combines the output candidates or results from both LiDAR and camera detection frameworks at a later stage. However, there is minimal emphasis on the quality of the training data, a crucial aspect in any detection framework. This issue is particularly evident in many fusion-based methods that lack sufficient sparse training samples for sparsely distributed object such as occluded and distant objects. The absence of comprehensive training data makes the trained model fragile and, as a result, incapable of effectively detecting distant objects during the testing phase. Consequently, fusion-based methods also face the challenge of inadequate data augmentation. The inherent disparities between 2D image data and 3D LiDAR data make it difficult to adapt several data augmentation techniques that are effective for the latter. This limitation poses a significant barrier and is a primary factor leading to the generally lower performance of multi-modal methods in comparison to their single-modal counterparts.

\subsection{Proposed Solution}
To address these issues, we propose  a simple yet effective ``VirtualPainting'' framework. Our method addresses the problem of sparse LiDAR points by generating virtual points using depth completion networks  PENET \cite{10105781}. To elaborate, we initiate the generation of supplementary virtual points and seamlessly merge them with the original points, resulting in an augmented LiDAR point cloud dataset. This augmented dataset subsequently undergoes a ``painting'' process utilizing features derived from cameras. The camera-derived features are in the form of semantic scores or per-pixel class scores. The augmented LiDAR point cloud is concatenated with per pixel class score to obtain a feature-rich point cloud. The result is twofold: it not only yields a denser point cloud in the form of augmented LiDAR point clouds but also enables a seamless combination of camera features and the point clouds. The virtual points, generated via the depth completion network, are now linked with camera features, resulting in a more comprehensive data representation. In certain scenarios where camera features were present but lacked corresponding lidar sensor points, these camera features remained unincorporated. Additionally, we address the challenges of insufficient training samples for distant objects and the absence of adequate data augmentation techniques by integrating a method called Distance Aware Data Augmentation (DADA). In this approach, we intentionally generate sparse training samples from objects that are initially densely observed by applying a distant offset. Considering that real-world scenes frequently involve incomplete data due to occlusion, we also introduce randomness by selectively removing portions to simulate such occlusion. By integrating these training samples, our model becomes more resilient during the testing phase, especially in the context of detecting sparsely distributed objects, such as occluded or distant objects, which are frequently overlooked in many scenarios. 

 In brief, our contributions are summarized as follows.
 \begin{itemize}
     \item An innovative approach that augments virtual points obtained through the joint application of LiDAR and Camera data with semantic labels derived from image-based semantic segmentation.
     \item Integration of distance-aware data augmentation method for improving models’ ability to identify sparse and occluded objects by creating training samples.
     \item A generic method that can incorporate any variation of 3D frameworks and 2D semantic segmentation for improving the overall detection accuracy.
     \item Evaluation on the KITTI and nuScenes dataset shows major improvement in both 3D and BEV detection benchmarks especially for distant objects.
 \end{itemize}

%-------------------------------------------------------------------------
\section{Related Work}
\begin{figure*}
  \centering
  \includegraphics[width=0.8\linewidth]{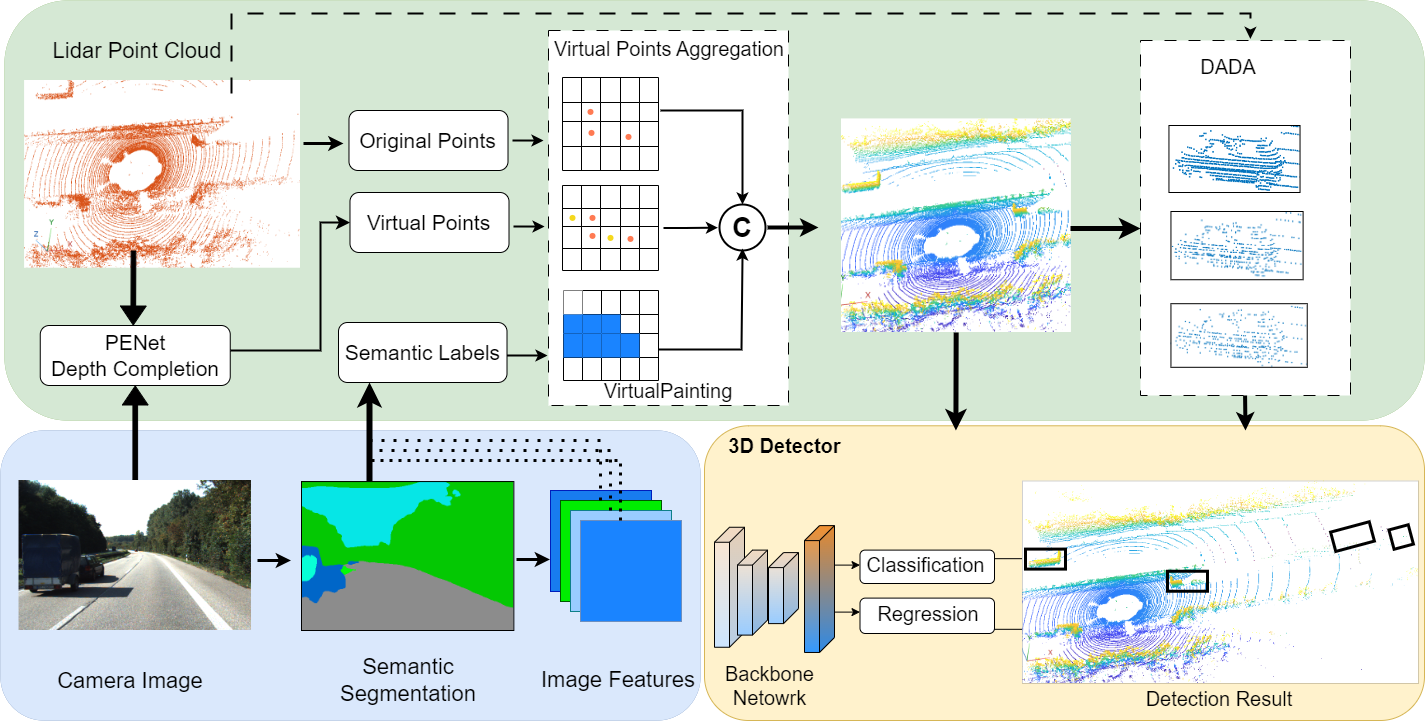}
  \caption{Overview of our VirtualPainting architecture which comprises five distinct stages: firstly, a 2D semantic segmentation module responsible for computing pixel-wise segmentation scores; secondly, an image-based depth completion network named "PENet" generates the LiDAR virtual point cloud; thirdly, the VirtualPainting process involves painting virtual and original LiDAR points with semantic segmentation scores; and fourthly, the Distance Aware Data Augmentation (DADA) component employs a distance-aware sampling strategy, creating sparse training samples primarily from nearby dense objects and finally a 3D detector for obtaining final detection result.}
  
\end{figure*}
\subsection{Single Modality}
Existing LiDAR-based methods can be categorized into four main groups based on their data representation: point-based, grid-based, point-voxel-based, and range-based. PointNet \cite{pointnet} and PointNet++ \cite{pointnetplus} are early pioneering works that apply neural networks directly to point clouds. PointRCNN \cite{pointrcnn}  introduced a novel approach, directly generating 3D proposals from raw point clouds. VoxelNet \cite{voxelnet} introduced the concept of a VFE (voxel feature encoding) layer to learn unified feature representations for 3D voxels. Building upon VoxelNet, CenterPoint \cite{centerpoint} devised an anchor-free method using a center-based framework based on CenterNet \cite{centernet}, achieving state-of-the-art performance. SECOND \cite{second} harnessed sparse convolution \cite{spconv} to alleviate the challenges of 3D convolution. PV-RCNN \cite{pvrcnn} bridged the gap between voxel-based and point-based methods to extract more discriminative features. Voxel-RCNN \cite{voxelrcnn} emphasized that precise positioning of raw points might not be necessary, contributing to the efficiency of 3D object detection.  PointPillars \cite{pointpillars} innovatively extracted features from vertical columns (Pillars) using PointNet \cite{pointnet}. Despite these advancements, all these approaches share a common challenge—the inherent sparsity of LiDAR point cloud data, which impacts their overall efficiency.

\subsection{Fusion-Based}
The inherent sparsity of LiDAR point cloud data has sparked research interest in multi-modal fusion-based methods. MV3D \cite{mv3d} and AVOD \cite{avod} create a multi-channel Bird's Eye View (BEV) image by projecting the raw point cloud into BEV. AVOD \cite{avod} takes both LiDAR point clouds and RGB images as input to generate features shared by the Region Proposal Network (RPN) and the refined network. MMF \cite{mmf} benefits from multi-task learning and multi-sensor fusion. 3D-CVF [2] fuses features from multi-view images, while Sniffer Faster RCNN \cite{snifferfasterrcnn} combines and refines the 2D and 3D proposal together at the final stage of detection. CLOCs \cite{clocs} and Sinffer Faster R-CNN++ \cite{10.1145/3631138} takes it one step further and refines the confidences of 3D candidates using 2D candidates in a learnable manner. These methods face limitations in utilizing image information due to the sparse correspondences between images and point clouds. Additionally, fusion-based methods encounter another challenge – a lack of sufficient data augmentation. In this paper, we address both issues by capitalizing on the virtual points and using data augmentation techniques.

\subsection{Point Decoration Fusion}
Recent developments in fusion-based methods have paved the way for innovative approaches like point decoration fusion-based methods, which enhance LiDAR points with camera features. PointPainting \cite{pointpainting}, for instance, suggests augmenting each LiDAR point with the semantic scores derived from camera images. PointAugmenting \cite{pointaugmenting} acknowledges the limitations of semantic scores and proposes enhancing LiDAR points with deep features extracted from a 2D object detectors. FusionPainting \cite{fusionpainting} takes a step further by harnessing both 2D and 3D semantic networks to extract additional features from both camera images and the LiDAR point cloud. Nevertheless, it's important to note that these methods also grapple with the sparse nature of LiDAR point cloud data as illustrated in Figure 1. Even though PointAugmenting addresses this issue with data augmentation techniques, the inadequacy of sparse training samples for sparse objects leads to their failure in detection.
%------------------------------------------------------------------------

\section{VirtualPainting}
We describe the architecture of our “VirtualPaintinag” framework in Fig. 2. To tackle the issue of sparsity of LiDAR point clouds and inadequate sparse training samples for these sparse objects, we introduce a multi-modality detector that enriches the original point cloud data through a series of different data enhancement processes.. As shown in Fig 2, the proposed framework consists of: (1) semantic segmentation module: an image based sem. seg. module that computes the pixel wise segmentation score. (2) Depth Completion Network: an image based depth completion network “PENet '' that generates the LiDAR virtual point cloud. (3) VirtualPainting: virtual and original LiDAR points are painted with sem. seg. scores. (4) Distance aware Data Augmentation(DADA): a distance aware sampling strategy that creates sparse training samples from nearby dense object   (4) 3D Detection Network: a LiDAR based 3D object detection network. 
\subsection{Image Based Semantic Network}
2D images captured by cameras are rich in texture, shape, and color information. This richness offers valuable complementary information for point clouds, ultimately enhancing 3D detection. To leverage this synergy, we employ a semantic segmentation network to generate pixel-wise semantic labels. We employ BiSeNet2 segmentation model for this purpose. This network takes multi-view images as its input and delivers pixel-wise classification labels for both foreground instances and the background. It's worth noting that our architecture is flexible, allowing for the incorporation of various semantic segmentation networks to generate semantic labels.

\subsection{PENet for Virtual Point Generation}
The geometry of nearby objects in LiDAR scans is often relatively complete, whereas for distant objects, it's quite the opposite. Additionally, there's a challenge of insufficient data augmentation due to the inherent disparities between 2D image data and 3D LiDAR data. Several data augmentation techniques that perform well with 3D LiDAR data are difficult to apply in multi-modal approaches. This obstacle significantly contributes to the usual under performance of multi-modal methods when compared to their single-modal counterparts. To address these issues, we employ PENet to transform 2D images into 3D virtual point clouds. This transformation unifies the representations of images and raw point clouds, allowing us to handle images much like raw point cloud data. We align the virtual points generated from the depth completion network with the original points to create a augmented point cloud data. This approach collectively enhances the geometric information of sparse objects while also establishing an environment for a unified representation of both images and point clouds.

\begin{figure}[t]
  \centering
  \includegraphics[width=1.0\linewidth]{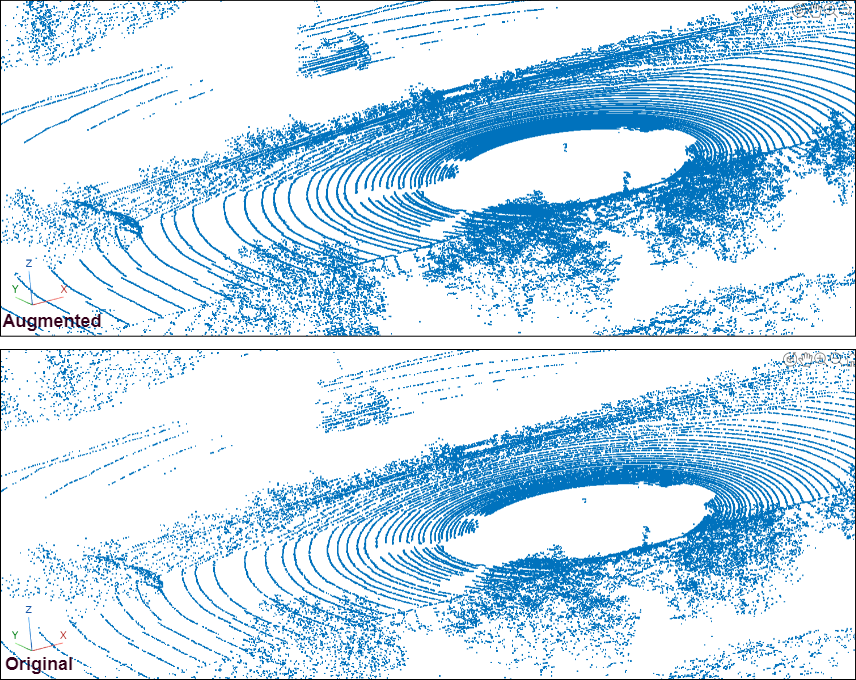}
  \caption{Illustration of augmented LiDAR point cloud and original LiDAR point cloud. The augmented LiDAR point cloud slightly more dense in comparison to the original, is the result of combining the original raw LiDAR points with the virtual points generated by PENet.}
  
\end{figure}

%-------------------------------

\subsection{Painting Virtual Points}
The current implementation of LiDAR point cloud painting methods has not harnessed the advantages of associating semantic results from the semantic network with virtual points. Incorporating semantic information into the augmented point cloud, which includes both the original and virtual points, not only enriches the dataset but also enhances the model's robustness. Let's refer the original points generated from LiDAR scan as "raw point cloud" deonted as $R$ and the points clouds generated from depth completion network as "virtual points" denoted by $V$. Starting with a set of raw clouds $R$, we have the capability to transform it into a sparse depth map $S$ using a known projection $T_{\text{LiDAR}\rightarrow\text{image}}$. We also have an associated image denoted as $I$ corresponding to $R$. By providing both $I$ and $S$ as inputs to a depth completion network, we obtain a densely populated depth map labeled as $D$. Utilizing a known projection $T_{\text{image}\rightarrow\text{LiDAR}}$, we can then generate a set of virtual points denoted as $V$.\\
 Our VirtualPainting algorithm comprises three primary stages. In the initial stage, utilizing the virtual points acquired from the depth completion network, we align these virtual points with the original raw LiDAR points, effectively generating an augmented LiDAR point cloud denoted as $A$ with N points. In the second stage, as previously mentioned, the segmentation network produces C-class scores. In the case of KITTI \cite{kitti}, C equals 4 (representing car, pedestrian, cyclist, and background), while for nuScenes \cite{nuscenes}, C is 11 (comprising 10 detection classes along with background). In the final stage, the augmented LiDAR points undergo projection onto the image, and the segmentation scores corresponding to the relevant pixel coordinates (h, w) are appended to the augmented LiDAR point, resulting in the creation of a painted LiDAR point. This transformation process involves a homogeneous transformation followed by projection into the image as given in Algorithm 1.
\begin{algorithm}[ht]
\caption{VirtualtPainting(L, V, A, S, T, M)}

\KwData{%
  \\
  LiDAR point cloud  $L \in \mathbb{R}^{N,D}$ with $N$ points and $D \geq 3$.\\
  Virtual Points  $V \in \mathbb{R}^{N,D}$ with $N$ points and $D \geq 3$.\\
  Augmented LiDAR point cloud A = L + V.
  Augmented LiDAR point cloud $A \in \mathbb{R}^{N,D}$ with $N$ points and $D \geq 3$.\\
  Segmentation scores $S \in \mathbb{R}^{W,H,C}$ with $C$ classes.\\
  Homogeneous transformation matrix $T \in \mathbb{R}^{4 \times 4}$.\\
  Camera matrix $M \in \mathbb{R}^{3 \times 4}$.
}
\KwResult{Painted augmented LiDAR points $P \in \mathbb{R}^{N,D+C}$}
\For{each $\tilde{l} \in A$}{
  $\tilde{l}_{\text{image}} = \text{PROJECT}(M, T, \tilde{l}_{\text{xyz}})$. \Comment{ $\tilde{l}_{\text{image}} \in \mathbb{R}^{2}$\\}
  
  $\tilde{s} = S[\tilde{l}_{\text{image}[0]}, \tilde{l}_{\text{image}[1]}, :]$ \Comment{$\tilde{s} \in \mathbb{R}^{C}$\\}
  
  $\tilde{p} = \text{Concatenate}(\tilde{l}, \tilde{s})$ \Comment{$\tilde{p} \in \mathbb{R}^{D+C}$}
}
\end{algorithm}
% \usepackage{booktabs}
% \usepackage{multirow}

\subsection{Distance Aware Data Augmentation}
As mentioned earlier, the absence of comprehensive geometric information for sparse objects can significantly hamper detection performance. To overcome this challenge, we aim to enhance our understanding of the geometry of sparsely observed distant objects by generating training samples derived from densely observed nearby objects. While several established methods exist to address this challenge, such as random sampling or farthest point sampling, it's important to note that these techniques often result in an uneven distribution pattern within the LiDAR-scanned point cloud. In this context, we adopt a sampling strategy \cite{ted} that takes into account both LiDAR scanning mechanics and scene occlusion. Within the context of a nearby ground truth box with position $C_g$ and its associated inside points $\{P_{gi}\}_i$, we introduce a random distance offset $\Delta\alpha$ as follows:
$C_{g} := C_{g} + \Delta\alpha, P_{gi} := P_{gi} + \Delta\alpha$.
Subsequently, we proceed to convert the points $\{P_{gi}\}_i$ into a spherical coordinate system and voxelize them into spherical voxels, aligning with the LiDAR's angular resolution. Within each voxel, we compute the distances between the points. If the points are found to be in very close proximity, with their distance being almost negligible, falling below a predefined threshold $\lambda$, we choose to calculate the average of these points. This results in a set of sampled points that closely mimic the distribution pattern of real scanned points, as depicted in Figure 2. During the training process, similar to the GT-AUG (Shi, Wang, and Li 2019) approach, we incorporate these sampled points and bounding box information into the training samples to facilitate data augmentation. This augmentation technique has the potential to address the shortage of training samples for distant objects. Additionally, we randomly remove portions of dense LiDAR points to simulate occlusion, aiming to potentially resolve the scarcity of occluded samples during training.
%DADA Algorithm 
%need to make some changes here
% \begin{algorithm}[H]
% 				\caption{Distance Aware Data Augmentation}
% 				\LinesNumbered
% 				\KwIn{LiDAR point cloud  $L \in \mathbb{R}^{N,D}$ with $N$ points and $D \geq 3$.\\}
% 				\KwOut{Processed LiDAR point cloud  $L_{P} \in \mathbb{R}^{M,D}$ with $M$ points and $D \geq 3$. ($M < N$)\\}
% 				\STATE  $x, y, z$ are vertex coordinates of each point in $L$, with $N$ points \\
% 				For each LiDAR point $p_i = (x_i, y_i, z_i) \in L$ \hfill \textbf{\\}
%   \Indp
%   %need to add actual algorithm here
%   \For{each $p_j = (x_j, y_j, z_j) \in L$ where $j \neq i$}{
%     \If{$(x_i - x_j, y_i - y_j, z_i - z_j) = (0, 0, 0)$}{
%       Output $L_i = (x_i, y_i, z_i)$
%     }
%   }
%   \Indm
				
% 				\Return Final Processed LiDAR Point Cloud $L_{p}
% 			\end{algorithm}
%------------------------------------------------------------------------
\section{3D Detector}
In the last phase of our architecture, the 3D detector receives the input in the form of the painted version of the augmented point cloud. As there are no alterations to the backbones or other architectural components, providing the painted point cloud as input to any 3D detector such as PointRCNN, VoxelNet, PVRCNN, PointPillars and so on, is very straightforward to obtain the final detection results. 
%------------------------------------------------------------------------
\section{Experimental Setup and Results}
We evaluate the effectiveness of our proposed "VirtualPainting" on both the large-scale autonomous driving 3D object detection KITTI and nuScences datasets. Initially, we provide an overview of our experimental setup, followed by presenting the evaluation results on both test and validation sets of each dataset.

% \usepackage{multirow}
\begin{table*}
\centering
\begin{tabular}{|l|l|l|lll|lll|lll|}
\hline
                         &                            & mAP                          & \multicolumn{3}{c|}{Car}                                                                                                             & \multicolumn{3}{c|}{Pedestrian}                                                                                                      & \multicolumn{3}{c|}{Cyclist}                                                                                                         \\ \cline{3-12} 
\multirow{-2}{*}{Method} & \multirow{-2}{*}{Modality} & Mod.                         & \multicolumn{1}{l|}{Easy}                         & \multicolumn{1}{l|}{Mod}                          & Hard                         & \multicolumn{1}{l|}{Easy}                         & \multicolumn{1}{l|}{Mod}                          & Hard                         & \multicolumn{1}{l|}{Easy}                         & \multicolumn{1}{l|}{Mod}                          & Hard                         \\ \hline
AVOD                     & L\&I                       & 64.07                        & \multicolumn{1}{l|}{90.99}                        & \multicolumn{1}{l|}{84.82}                        & 79.62                        & \multicolumn{1}{l|}{58.49}                        & \multicolumn{1}{l|}{50.32}                        & 46.98                        & \multicolumn{1}{l|}{69.39}                        & \multicolumn{1}{l|}{57.12}                        & 51.09                        \\ \hline
MV3D                     & L\&I                       & N/A                          & \multicolumn{1}{l|}{86.62}                        & \multicolumn{1}{l|}{78.93}                        & 69.80                        & \multicolumn{1}{l|}{N/A}                          & \multicolumn{1}{l|}{N/A}                          & N/A                          & \multicolumn{1}{l|}{N/A}                          & \multicolumn{1}{l|}{N/A}                          & N/A                          \\ \hline
PointRCNN                & L                      & 66.92                        & \multicolumn{1}{l|}{92.13}                        & \multicolumn{1}{l|}{87.39}                        & 82.72                        & \multicolumn{1}{l|}{54.77}                        & \multicolumn{1}{l|}{46.13}                        & 42.84                        & \multicolumn{1}{l|}{82.56}                        & \multicolumn{1}{l|}{67.24}                        & 60.28                        \\ \hline
VirtualPointRCNN         & L\&I                       & 70.56                        & \multicolumn{1}{l|}{92.67}                        & \multicolumn{1}{l|}{88.35}                        & 83.67                        & \multicolumn{1}{l|}{58.99}                        & \multicolumn{1}{l|}{50.22}                        & 46.63                        & \multicolumn{1}{l|}{84.07}                        & \multicolumn{1}{l|}{71.76}                        & 63.17                        \\ \hline
Improvement ↑                   &                            & {\color[HTML]{009901} +3.64} & \multicolumn{1}{l|}{{\color[HTML]{009901} +0.54}} & \multicolumn{1}{l|}{{\color[HTML]{009901} +0.96}} & {\color[HTML]{009901} +0.95} & \multicolumn{1}{l|}{{\color[HTML]{009901} +4.22}} & \multicolumn{1}{l|}{{\color[HTML]{009901} +4.09}} & {\color[HTML]{009901} +3.79} & \multicolumn{1}{l|}{{\color[HTML]{009901} +1.51}} & \multicolumn{1}{l|}{{\color[HTML]{009901} +4.52}} & {\color[HTML]{009901} +2.89} \\ \hline
PointPillars             & L                          & 65.98                        & \multicolumn{1}{l|}{90.07}                        & \multicolumn{1}{l|}{86.56}                        & 82.81                        & \multicolumn{1}{l|}{57.60}                        & \multicolumn{1}{l|}{48.64}                        & 45.78                        & \multicolumn{1}{l|}{79.90}                        & \multicolumn{1}{l|}{62.73}                        & 55.58                        \\ \hline
VirtualPointPillars      & L\&I                       & 69.07                        & \multicolumn{1}{l|}{90.39}                        & \multicolumn{1}{l|}{86.98}                        & 83.59                        & \multicolumn{1}{l|}{58.86}                        & \multicolumn{1}{l|}{50.03}                        & 46.71                        & \multicolumn{1}{l|}{80.15}                        & \multicolumn{1}{l|}{65.89}                        & 59.44                        \\ \hline
Improvement ↑                   &                            & {\color[HTML]{009901} +3.09} & \multicolumn{1}{l|}{{\color[HTML]{009901} +0.32}} & \multicolumn{1}{l|}{{\color[HTML]{009901} +0.42}} & {\color[HTML]{009901} +0.78} & \multicolumn{1}{l|}{{\color[HTML]{009901} +1.26}} & \multicolumn{1}{l|}{{\color[HTML]{009901} +1.39}} & {\color[HTML]{009901} +0.93} & \multicolumn{1}{l|}{{\color[HTML]{009901} +0.25}} & \multicolumn{1}{l|}{{\color[HTML]{009901} +3.16}} & {\color[HTML]{009901} +3.86} \\ \hline
PVRCNN                   & L                          & 68.54                        & \multicolumn{1}{l|}{91.91}                        & \multicolumn{1}{l|}{88.13}                        & 85.40                        & \multicolumn{1}{l|}{52.41}                        & \multicolumn{1}{l|}{44.83}                        & 42.57                        & \multicolumn{1}{l|}{79.60}                        & \multicolumn{1}{l|}{64.46}                        & 57.94                        \\ \hline
VirtualPVRCNN            & L\&I                       & 70.05                        & \multicolumn{1}{l|}{92.33}                        & \multicolumn{1}{l|}{88.64}                        & 86.08                        & \multicolumn{1}{l|}{53.57}                        & \multicolumn{1}{l|}{48.97}                        & 46.76                        & \multicolumn{1}{l|}{80.26}                        & \multicolumn{1}{l|}{68.39}                        & 61.91                        \\ \hline
Improvement ↑                   &                            & {\color[HTML]{009901} +1.51} & \multicolumn{1}{l|}{{\color[HTML]{009901} +0.42}} & \multicolumn{1}{l|}{{\color[HTML]{009901} +0.51}} & {\color[HTML]{009901} +0.68} & \multicolumn{1}{l|}{{\color[HTML]{009901} +1.16}} & \multicolumn{1}{l|}{{\color[HTML]{009901} +4.14}} & {\color[HTML]{009901} +4.19} & \multicolumn{1}{l|}{{\color[HTML]{009901} +0.66}} & \multicolumn{1}{l|}{{\color[HTML]{009901} +3.93}} & {\color[HTML]{009901} +3.97} \\ \hline
\end{tabular}
\caption{Results on the KITTI test BEV detection benchmark. All three baseline models exhibit enhancements across all levels within the three classes. This improvement is particularly pronounced in classes like pedestrian and cyclist, where there is a greater likelihood of remaining undetected when positioned at a greater distance.}
\end{table*}
\subsection{Semantic Network Details}
For semantic segmentation in our KITTI dataset experiments, we use the BiSeNetv2 \cite{bisenetv2} network. This network underwent an initial pretraining phase on the CityScapes \cite{cityscapes} dataset and was subsequently fine-tuned on the KITTI dataset using PyTorch \cite{pytorch}. For the simplicity of our implementation, we chose to ignore classes cyclist and bike because there is a difference in the class defination of a cyclist between KITTI semantic segmentation and object detection tasks. In object detection, a cyclist is defined as a combination of the rider and the bike, whereas in semantic segmentation, a cyclist is defined as solely the rider, with the bike being a separate class.
Similiary for nuScenes, we developed a custom network using the nuImages dataset, which comprises 100,000 images containing 2D bounding boxes and segmentation labels for all nuScenes classes. The segmentation network utilized a ResNet backbone to extract features at various strides, ranging from 8 to 64, and incorporated an FCN segmentation head for predicting nuScenes segmentation scores.
\subsection{PENet}
Much like SFD, our approach relies on the virtual points produced by PENet. The  PENet architecture is initially trained exclusively on the KITTI dataset, which encompasses both color images and aligned sparse depth maps. These depth maps are created by projecting 3D LiDAR points onto their corresponding image frames. The dataset comprises 86,000 frames designated for training, in addition to 7,000 frames allocated for validation, and a further 1,000 frames designated for testing purposes. Our PENEt is trained on the training set.The images are standardized at a resolution of 1216 × 352. Typically, a sparse depth map contains approximately 5 valid pixels, while ground truth dense depth maps have an approximate 16 coverage of valid pixels \cite{9880017}.
%https://arxiv.org/pdf/2103.00783.pdf PENET
\subsection{LiDAR Network Details}
We use the OPENPCDet \cite{openpcdet2020} tool for the KITTI dataset, incorporating 3D detectors such as PVRCNN, PointPillars, PointRCNN, and VoxelNEt, with only minimal adjustments to the point cloud dimension. In order to accommodate segmentation scores from a semantic network, we expand the dimension by adding the total number of classes used in the segmentation network. Since our architecture remains unchanged beyond this point, it remains generic and can be readily applied to any 3D detector without requiring complex configuration modifications. Similarly, for the nuScenes dataset, we utilize an enhanced version of PointPillars, as detailed in \cite{pointpainting}. These enhancements involve alterations to the pillar resolution, network architecture, and data augmentation techniques. Firstly, we reduce the pillar resolution from 0.25 meters to 0.2 meters to improve the localization of small objects. Secondly, we revise the network architecture to incorporate additional layers earlier in the network. Lastly, we adjust the global yaw augmentation from $\pi$ to $\frac{\pi}{6}$ \cite{pointpainting}.
%need to check this section again

\subsection{KITTI Results}
We initially assess our model's performance using the KITTI dataset and contrast it with the current state-of-the-art methods. In Table 1, you can observe the outcomes for the KITTI test BEV detection benchmark. Notably, there is a substantial enhancement in mean average precision (mAP) compared to both single and multi-modality baseline methods like AVOD, MV3D, PointRCNN, PointPillars, and PVRCNN. This improvement is particularly prominent in the pedestrian and cyclist categories, which are more challenging to detect, especially when it comes to sparse objects. Notably, the moderate and hard classes within the pedestrian and cyclist categories exhibit more significant improvements when compared to the easy class, as depicted in Table 1. We can clearly observe that for PointRCNN, our approach exhibits a notable enhancement of \textcolor[HTML]{009901}{+4.09, +3.79 and +4.52, +2.89}, specifically for the moderate and hard difficulty levels within the pedestrian and cyclist classes. This trend is consistent across various other models as well. Similarly, Table 2 provides a comparison of our methods with point-painting based approaches. Although there isn't a remarkable increase in mean average precision, there is still some notable enhancement.

\begin{table*}
\centering

\begin{tabular}{|l|l|lll|l|l|}
\hline
\multicolumn{1}{|c|}{}                        & \multicolumn{1}{c|}{}                           & \multicolumn{3}{c|}{mAP}                                                                    & \multicolumn{1}{c|}{}                                                                      & \multicolumn{1}{c|}{}                              \\ \cline{3-5}
\multicolumn{1}{|c|}{\multirow{-2}{*}{Model}} & \multicolumn{1}{c|}{\multirow{-2}{*}{Modality}} & \multicolumn{1}{c|}{Car}   & \multicolumn{1}{c|}{Pedestrian} & \multicolumn{1}{c|}{Cyclist} & \multicolumn{1}{c|}{\multirow{-2}{*}{\begin{tabular}[c]{@{}c@{}}Total\\ mAP\end{tabular}}} & \multicolumn{1}{c|}{\multirow{-2}{*}{Improvement}} \\ \hline
PaintedPointRCNN                              & L\&I                                            & \multicolumn{1}{c|}{87.97} & \multicolumn{1}{c|}{51.97}      & \multicolumn{1}{c|}{72.80}   & \multicolumn{1}{c|}{70.91}                                                                 & \multicolumn{1}{c|}{}                              \\ \hline
VirtualPointRCNN                              & L\&I                                            & \multicolumn{1}{c|}{88.23} & \multicolumn{1}{c|}{52.61}      & \multicolumn{1}{c|}{72.66}   & \multicolumn{1}{c|}{71.16}                                                                 & {\color[HTML]{009901} +0.25}                       \\ \hline
PaintedPointPillars                           & L\&I                                            & \multicolumn{1}{l|}{87.18} & \multicolumn{1}{l|}{51.44}      & 68.18                        & 69.02                                                                                      &                                                    \\ \hline
VirtualPointPillars                           & L\&I                                            & \multicolumn{1}{l|}{87.32} & \multicolumn{1}{l|}{51.87}      & 68.49                        & 69.23                                                                                      & {\color[HTML]{009901} +0.21}                       \\ \hline
PaintedPVRCNN                                 & L\&I                                            & \multicolumn{1}{l|}{88.91} & \multicolumn{1}{l|}{48.78}      & 69.97                        & 69.22                                                                                      &                                                    \\ \hline
VirtualPVRCNN                                 & L\&I                                            & \multicolumn{1}{l|}{89.35} & \multicolumn{1}{l|}{49.43}      & 70.08                        & 69.98                                                                                      & {\color[HTML]{009901} +0.76}                       \\ \hline
\end{tabular}

\caption{Comparision of PointPainting based models with our VirtualPainting based models.}
\end{table*}
\begin{table*}
\small
\centering 
\begin{tabular}{lccccccccccccl}
\hline
\multicolumn{1}{c|}{}                         & \multicolumn{1}{l|}{}                           & \multicolumn{1}{l|}{}                      & \multicolumn{1}{l|}{}                      & \multicolumn{10}{c}{AP}                                                                                                                                                                                                                                                                                            \\ \cline{5-14} 
\multicolumn{1}{c|}{\multirow{-2}{*}{Method}} & \multicolumn{1}{l|}{\multirow{-2}{*}{Modality}} & \multicolumn{1}{l|}{\multirow{-2}{*}{NDS}} & \multicolumn{1}{l|}{\multirow{-2}{*}{mAP}} & \multicolumn{1}{l|}{Car}    & \multicolumn{1}{l|}{Truck}   & \multicolumn{1}{l|}{Bus}     & \multicolumn{1}{l|}{Trailer} & \multicolumn{1}{l|}{Cons}    & \multicolumn{1}{l|}{Ped}     & \multicolumn{1}{l|}{Moto}    & \multicolumn{1}{l|}{Bicycle} & \multicolumn{1}{l|}{TC}      & \multicolumn{1}{l}{Barrier} \\ \hline
3DCVF                         & L \& C                                          & 62.3                                       & 52.7                                       & 83.0                        & 45.0                         & 48.8                         & 49.6                         & 15.9                         & 74.2                         & 51.2                         & 30.4                         & 62.9                         & 65.9                         \\
PointPillars                 & L                                               & 55.0                                       & 40.1                                       & 76.0                        & 31.0                         & 32.1                         & 36.6                         & 11.3                         & 64.0                         & 34.2                         & 14.0                         & 45.6                         & 56.4                         \\
PaintedPointPillars            & L \& C                                          & 58.1                                       & 46.4                                       & 77.9                        & 35.8                         & 36.2                         & 37.3                         & 15.8                         & 73.3                         & 41.5                         & 24.1                         & 62.4                         & 60.2                         \\
VirtualPointPillars                           & L \& C                                          & 60.21                                      & 48.59                                      & 79.20                       & 36.11                        & 39.47                        & 40.53                        & 16.27                        & 76.25                        & 47.45                        & 28.79                        & 65.71                        & 63.77                        \\
Improvement ↑                                 & -                                        & {\color[HTML]{009901} +2.11}               & {\color[HTML]{009901} +2.19}               & {\color[HTML]{009901} +1.3} & {\color[HTML]{009901} +0.29} & {\color[HTML]{009901} +3.07} & {\color[HTML]{009901} +3.23} & {\color[HTML]{009901} +0.47} & {\color[HTML]{009901} +2.95} & {\color[HTML]{009901} +5.95} & {\color[HTML]{009901} +4.69} & {\color[HTML]{009901} +3.31} & {\color[HTML]{009901} +3.57} \\ \hline
\end{tabular}
\caption{Comparisons of performance on the nuScenes test set, with reported metrics including NDS, mAP, and class-specific AP.}
\end{table*}

\begin{table}[]
\centering
\resizebox{\columnwidth}{!}{%
\begin{tabular}{ccccc}
\hline
\multicolumn{1}{c|}{VirtualPainting} & \multicolumn{1}{c|}{PointPainting} & \multicolumn{1}{c|}{F-PointNet} & \multicolumn{1}{c|}{EPNet} & \multicolumn{1}{c}{3D-CVF} \\ \hline \hline
5.37FPS                                & 2.38FPS                             & 5.52FPS                          & 8.15FPS                      & 13.83FPS \\ \hline                
\end{tabular}%
}
\caption{Inference speed across various multi and single modality frameworks}
\end{table}

\subsection{nuScenes Results}
Additionally, we assess the performance of our model using the nuScenes dataset, as displayed in Table 3. Our approach surpasses the single modality PointPillar-based model in all categories. Furthermore, the enhanced variant called PaintedPointPillars, which is based on the point-painting method for PointPillars, exhibits improvements in terms of nuScenes Detection Score (NDS) and Average Precision (AP) across all ten classes in the nuScenes dataset. Similarly, as seen in our previous results, this improvement is most noticeable in classes such as Pedestrian, Bicycle, and Motorcycle, where the likelihood of remaining undetected in occluded and distant sparse regions is higher.
%------------------------------------------------------------------------
%

%nuScenes result

\section{Ablation Studies}
%KITTI dataset
% \usepackage{multirow}
% \usepackage{graphicx}
\begin{table}[]
\centering
\resizebox{\columnwidth}{!}{%
\begin{tabular}{c|ccc|cc}
\hline
\multirow{2}{*}{Model}                  & \multirow{2}{*}{VPC} & \multirow{2}{*}{Sem.Seg} & \multirow{2}{*}{DA-Aug.} & \multicolumn{2}{c}{3D AP}                 \\ \cline{5-6} 
                                        &                      &                          &                          & \multicolumn{1}{c|}{Pedestrian} & Cyclist \\ \hline \hline
\multicolumn{1}{l|}{PointRCNN}          & \multicolumn{3}{c|}{}                                                      & \multicolumn{1}{c|}{47.85}      & 68.93   \\
\multicolumn{1}{l|}{VirtualPointRCNN}   & \checkmark                    &                          & \checkmark                        & \multicolumn{1}{c|}{48.43}      & 69.39   \\
                                        & \checkmark                    & \checkmark                         &                          & \multicolumn{1}{c|}{51.75}      & 72.04   \\
                                        & \checkmark                    & \checkmark                         & \checkmark                         & \multicolumn{1}{c|}{52.61}      & 72.66   \\ \hline \hline
\multicolumn{1}{l|}{PointPillar}        & \multicolumn{3}{c|}{}                                                      & \multicolumn{1}{c|}{50.13}      & 66.16   \\
\multicolumn{1}{l|}{VirtualPointPillar} & \checkmark                     &                          & \checkmark                         & \multicolumn{1}{c|}{50.59}      & 66.62   \\
                                        & \checkmark                     & \checkmark                         &                          & \multicolumn{1}{c|}{51.45}      & 68.06   \\
                                        & \checkmark                     & \checkmark                         & \checkmark                         & \multicolumn{1}{c|}{51.87}      & 68.49   \\ \hline \hline
\multicolumn{1}{l|}{PVRCNN}             & \multicolumn{3}{c|}{}                                                      & \multicolumn{1}{c|}{46.41}      & 67.26   \\
\multicolumn{1}{l|}{VirtualPVRRCNN}     & \checkmark                     &                          & \checkmark                         & \multicolumn{1}{c|}{46.83}      & 67.73   \\
                                        & \checkmark                     & \checkmark                         &                          & \multicolumn{1}{c|}{48.94}      & 69.48   \\
                                        & \checkmark                     & \checkmark                         & \checkmark                         & \multicolumn{1}{c|}{49.43}      & 70.08   \\ \hline
\end{tabular}%
}
\caption{Analysis of the impacts of different components on the KITTI test set, with results assessed using the AP (Average Precision) metric calculated across 40 recall positions specifically for pedestrian and cyclist class.}
\end{table}

%nuscenes dataset

% \usepackage{graphicx}
% \usepackage{graphicx}
% \usepackage{graphicx}
\begin{table}[]
\centering
\resizebox{\columnwidth}{!}{%
\begin{tabular}{l|ccccc}
\hline
    & \multicolumn{1}{l|}{VPC}  & \multicolumn{1}{l|}{Sem.Seg} & \multicolumn{1}{l|}{DADA} & \multicolumn{1}{l|}{mAP} & \multicolumn{1}{l}{NDS} \\ \hline \hline
(a) &                           &                              &                            & \multicolumn{1}{l}{40.12} & \multicolumn{1}{l}{55.02}                   \\
(b) & \checkmark &                              & \checkmark & 41.03 {\small \color[HTML]{009901}+0.93}              & 55.75 {\small \color[HTML]{009901}+0.75}                 \\
(c) &                           & \checkmark    & \checkmark & 47.05 {\small \color[HTML]{009901}+6.75}
             & 58.73 {\small \color[HTML]{009901}+2.98 }                 \\
(d) & \checkmark & \checkmark    &                           & 47.78 {\small \color[HTML]{009901}+0.73}              & 59.59 {\small \color[HTML]{009901}+0.86 }                 \\
(e) & \checkmark & \checkmark    & \checkmark & 48.59 {\small \color[HTML]{009901}+0.81}             & 60.21 {\small \color[HTML]{009901}+0.62 }                  \\ \hline
\end{tabular}%
}
\caption{Ablation Studies for different component on nuScenes dataset.Similar to KITTI, the semantic network plays a pivotal role in enhancing performance in the nuScenes dataset.}
\end{table}

\subsection{VirtualPainting is a generic and flexible method}
As depicted in Table 1 and Table 2, we assess the genericity of our approach by integrating it with established 3D detection frameworks. We carry out three sets of comparisons, each involving a single-modal method and its multi-modal counterpart. The three LiDAR-only models under consideration are PointPillars, PointRCNN, and PVRCNN. As illustrated in Table 2, VirtualPainting consistently demonstrates enhancements across all single-modal detection baselines. These findings suggest that VirtualPainting possesses a general applicability and could potentially be extended to other 3D object detection frameworks. Likewise, Table X demonstrates that several elements, including virtual points and semantic segmentation, can be seamlessly incorporated into the architecture without requiring intricate modifications, thus highlighting the flexibility of our approach. Similary, we also evaluated the inference speed of our method on an NVIDIA RTX 3090 GPU. While the inference speed is higher compared to the original PointPainting method, it still remains faster than other single-modality based methods, as indicated in Table 4.

\subsection{Where does the improvement come from?}
\begin{figure}[t]
  \centering
  \includegraphics[width=0.8\linewidth]{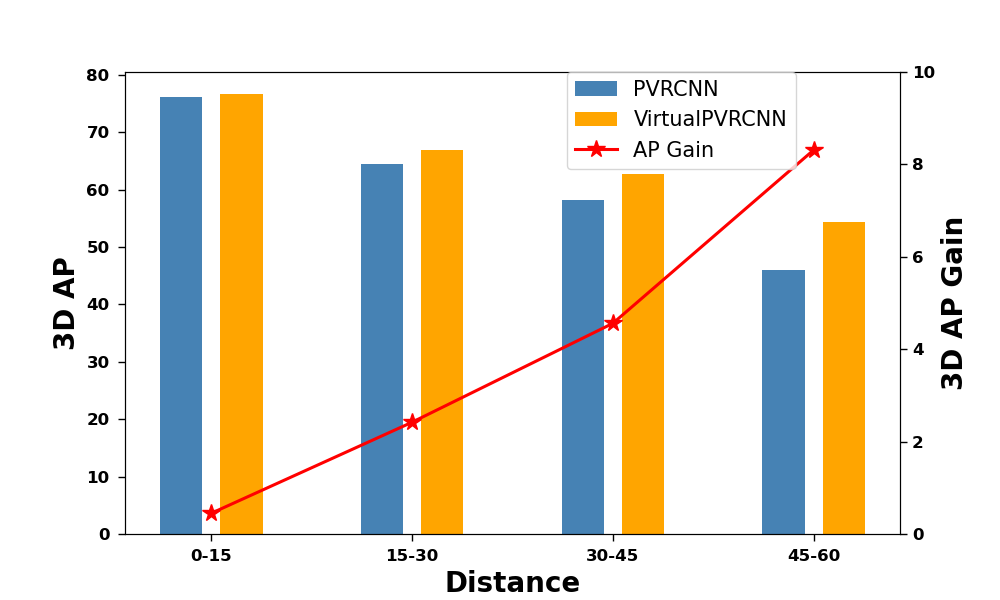}
  \caption{Performance improvement along different detection distance (KITTI test set) and 3D AP for cyclist class.}
 
\end{figure}

\begin{figure}[t]
  \centering
  \includegraphics[width=0.8\linewidth]{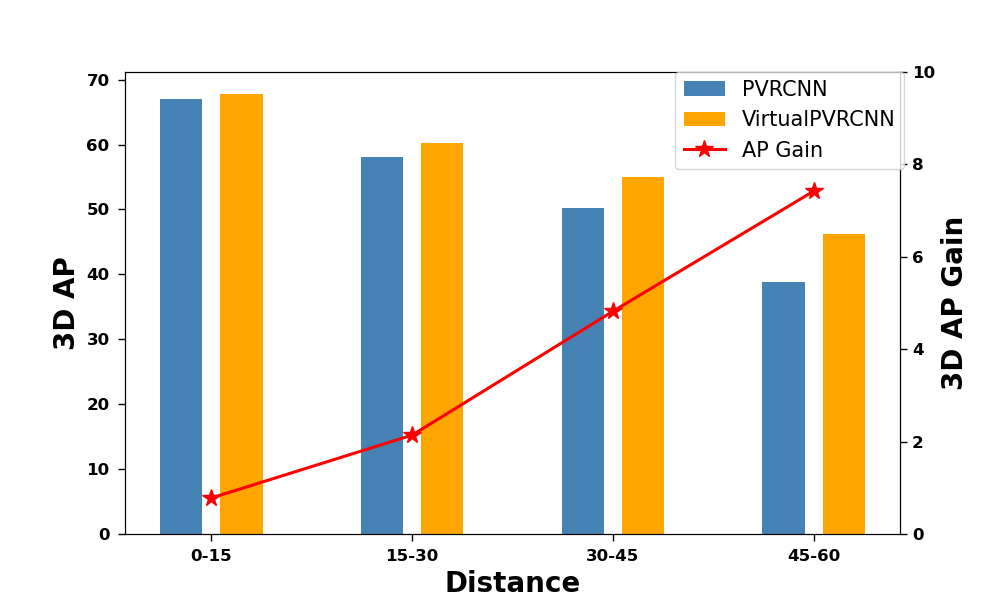}
  \caption{Performance improvement along different detection distance (KITTI test set) and 3D AP for pedestrian class.}
  
\end{figure}

To gain a comprehensive understanding of how VirtualPainting leverages camera cues and augmentation techniques applied to LiDAR points to enhance 3D object detection models, we offer a thorough analysis encompassing both qualitative and quantitative aspects. Initially, we categorize objects into three groups based on their proximity to the ego-car: those within 30 meters, those falling within the 30 to 50-meter range, and those located beyond 50 meters.Figure 4 and Figure 5 illustrates the relative improvements achieved through multi-modal fusion within each of these distance groups. In essence, VirtualPainting consistently enhances accuracy across all distance ranges. Notably, it delivers significantly higher accuracy gains for long-range objects compared to short-range ones (e.g). This phenomenon may be attributed to the fact that long-range objects often exhibit sparse LiDAR point coverage, and the combination of augmentation techniques applied to these sparse points, rendering them denser, along with the inclusion of high-resolution camera semantic labels, effectively bridges the information gap. Likewise, as illustrated in Table 5 and Table 6, the incorporation of semantic segmentation significantly elevates precision compared to other elements, including the virtual point cloud and DADA. Although there is noticeable improvement when both virtual points and DADA are integrated, the primary contribution stems from the attachment of semantic labels to the augmented point cloud, which contains both virtual points and the original LiDAR points for both KITTI and nuScenes datasets.
% \usepackage{graphicx}

%probably add some real word experiments

\section{Conclusion}
In this paper, we propose a generic 3D object detector that combines both LiDAR point cloud and camera image to improve the detection accuracy, specially for sparsely distributed objects. We address the  sparsity and inadequate data augmentation problems of LiDAR point cloud through the combined application of camera and LiDAR data. Using depth completion network we generate virtual point cloud to make the LiDAR points cloud dense and adopt the point decoration mechanism to decorate the augmented LiDAR points cloud with image semantics thus improving the detection for those objects that generally goes undetected due to sparse LiDAR points. Besides, we design a distance aware data augmentation techniques to make the model robust for occluded and distant object. Experimental results demonstrate that our approach can significantly improve detection accuracy, particularly for these specific objects.

%%%%%%%%% REFERENCES
{\small
\bibliographystyle{ieee_fullname}
\bibliography{references}
}

\end{document}